\documentclass{article}
\usepackage{color,xcolor}
\usepackage{epsfig}
\usepackage{graphicx}
\usepackage{subfiles}

\usepackage{adjustbox}
\usepackage{array}
\usepackage{booktabs}
\usepackage{colortbl}
\usepackage{float,wrapfig}
\usepackage{hhline}
\usepackage{multirow}
\usepackage{subcaption} 
\usepackage[labelfont={bf},labelsep={period},font={small}]{caption}

\let\llncssubparagraph\subparagraph
\let\subparagraph\paragraph
\usepackage[compact]{titlesec}
\let\subparagraph\llncssubparagraph

\usepackage{amsmath,amsfonts,amssymb}
\usepackage{bm}
\usepackage{nicefrac}

\usepackage{changepage}
\usepackage{extramarks}
\usepackage{fancyhdr}
\usepackage{lastpage}
\usepackage{setspace}
\usepackage{soul}
\usepackage{xspace}

\usepackage{url}
\usepackage[numbers]{natbib}
\usepackage{algorithm}
\usepackage{algpseudocode}
\usepackage{enumerate}

\usepackage{times}

\usepackage[bookmarks=false]{hyperref}

\usepackage{pbox}
\usepackage{footnote}
\usepackage{tablefootnote}

\usepackage{paralist}
\usepackage{paralist}

\newcommand{\myparagraph}[1]{\vspace{-5pt}\paragraph{#1}}

\makeatletter
\DeclareRobustCommand\onedot{\futurelet\@let@token\@onedot}
\def\@onedot{\ifx\@let@token.\else.\null\fi\xspace}

\def\eg{\emph{e.g}\onedot}

\def\etc{\emph{etc}\onedot} \def\vs{\emph{vs}\onedot}

\makeatother

\newcommand{\itddd}{I3D\textsubscript{3$\times$3$\times$3}\xspace}
\newcommand{\itd}{I3D\textsubscript{3$\times$1$\times$1}\xspace}
\newcommand{\tsm}{TSM\textsubscript{8f}\xspace}

\makeatletter
\newcommand\footnoteref[1]{\protected@xdef\@thefnmark{\ref{#1}}\@footnotemark}
\makeatother
 
     \usepackage[preprint]{neurips_2019}

\title{Training Kinetics in 15 Minutes: \\  
Large-scale Distributed Training on Videos}

\author{%
  Ji Lin \\
  MIT \\
  \texttt{jilin@mit.edu} \\
   \And
   Chuang Gan \\
   MIT-IBM Watson AI Lab \\
   \texttt{ganchuang@csail.mit.edu} \\
   \And
   Song Han \\
   MIT \\
   \texttt{songhan@mit.edu} \\
}

\begin{document}

\maketitle
\begin{abstract}
    Deep video recognition is more computationally expensive than image recognition, especially on large-scale datasets like Kinetics~\cite{carreira2017quo}. Therefore, training scalability is essential to handle a large amount of videos.
    In this paper, we study the factors that impact the training scalability of video networks. We recognize three bottlenecks, including data loading (data movement from disk to GPU), communication (data movement over networking), and computation FLOPs. We propose three design guidelines to improve the scalability: (1) fewer FLOPs and hardware-friendly operator to increase the \emph{computation efficiency}; (2) fewer input frames to reduce the data movement and increase the \emph{data loading efficiency}; (3) smaller model size to reduce the networking traffic and increase the \emph{networking efficiency}. With these guidelines, we designed a new operator  Temporal Shift Module (TSM) that is efficient and scalable for distributed training. TSM model can achieve $1.8\times$ higher throughput compared to previous I3D models. We scale up the training of the TSM model to 1,536 GPUs, with a mini-batch of 12,288 video clips/98,304 images, without losing the accuracy. With such hardware-aware model design, we are able to scale up the training on Summit supercomputer and reduce the training time on Kinetics dataset from 49 hours 55 minutes to 14 minutes 13 seconds, achieving a top-1 accuracy of 74.0\%, which is 1.6$\times$ and 2.9$\times$ faster than previous 3D video models with higher accuracy. 
    The code and more details can be found here: \textcolor{purple}{\url{http://tsm-hanlab.mit.edu}}.
\end{abstract}
\section{Introduction}

Videos are increasing explosively in various areas, including healthcare, self-driving cars, virtual reality, \etc. To handle the massive amount of videos collected everyday, we need a scalable video training system to enable fast learning. Existing works on distributed training mostly focus on image recognition~\cite{goyal2017accurate, you2017scaling, you2018imagenet,chen2016revisiting, smith2017don, you2019large,you2017large}. For video recognition, the problem is more challenging but less explored: (1) the video models usually consume an order of magnitude larger computation compared to the 2D image counterpart. For example, a widely used ResNet-50~\cite{he2016deep} model has around 4G FLOPs, while a ResNet-50 I3D~\cite{wang2017non} consumes 33G FLOPs, more than 8$\times$ larger; (2) video datasets are far more larger than 2D image datasets, and the data I/O is much higher than images. For example, ImageNet~\cite{russakovsky2015imagenet} has 1.28M training images, while a video dataset Kinetics-400 has 63M training frames, which is around 50$\times$ larger; (3) video models are usually larger in model size, thus it requires higher networking bandwidth to exchange the gradients.


In this paper, we study the bottlenecks of large-scale distributed training on videos, including computation, data loading (I/O), and communication. Correspondingly, we propose three practical design guidelines to solve the challenges: the model should leverage hardware-friendly operators to reduce the computation FLOPs; the model should take fewer input frames to save the file system I/O; the model should use operators with fewer parameters to save the networking bandwidth.
Under such guidelines, we propose an efficient video CNN operator: the Temporal Shift Module (TSM) that has \emph{zero FLOPs and zero parameters}. It can scale up the Kinetics training to 1,536 GPUs, reaching a mini-batch size of 12,288 video clips/98,304 images. The whole training process can finish within 15 minutes and achieve a top-1 accuracy of 74.0\%. Compared to previous I3D models~\cite{hara2018can, wang2017non}, our TSM model can increase the training speed by 1.6$\times$ and 2.9$\times$ on the world-leading supercomputer Summit.

\section{Preliminary}

We first introduce the preliminary knowledge about video recognition networks. Different from 2D image recognition, the input/activation of a video network has 5 dimensions $[N, T, C, H, W]$, where $N$ is the batch size, $T$ is the temporal timestamps, $C$ is the channel number, $H$ and $W$ are the spatial resolution. A simple video recognition model directly applys 2D CNN to each frames~\cite{wang2016temporal, karpathy2014large, simonyan2014two}. However, such methods cannot model the temporal relationships between frames, which are necessary for video understanding~\cite{lin2018tsm, wang2017non}. To perform spatial-temporal feature learning, an effective way is to inflate 2D convolutions into 3D convolutions. The resulting model is I3D~\cite{carreira2017quo, tran2015learning}. However, by inflating temporal dimension by $k$, the computation and the model size is also increased by $k$ times.
Given such dilemma, we study two aspects of model design to make it more hardware efficient:


\begin{figure}[t]
\centering
\includegraphics[width=0.8\textwidth]{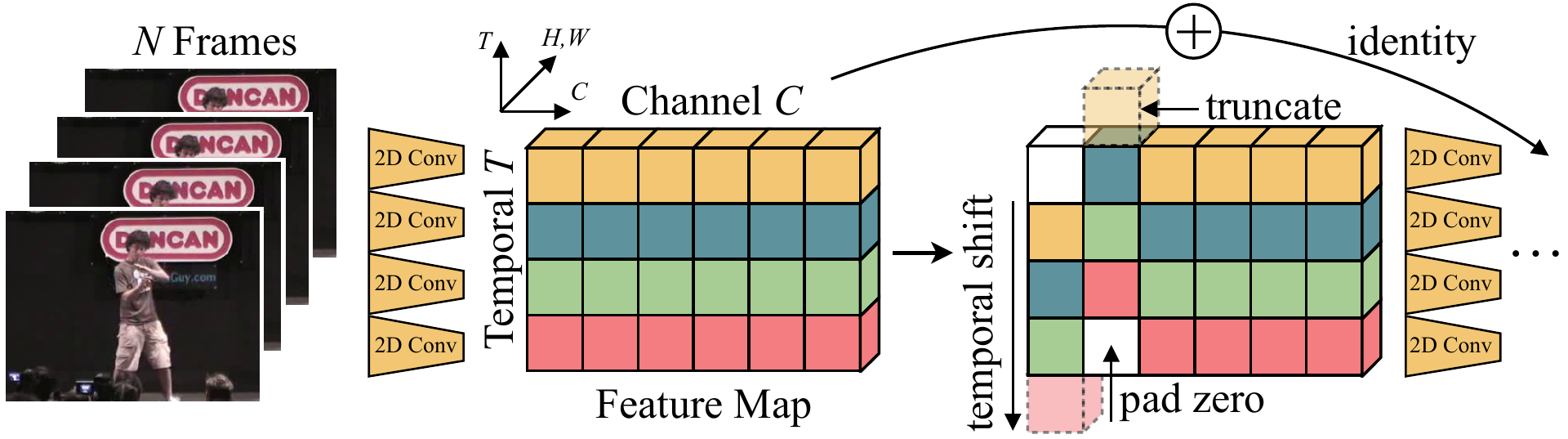}
\caption{Temporal Shift Module (TSM) shifts channels along temporal dimension to enable temporal modeling among neighboring frames. It has \emph{zero FLOPs and zero parameters}. It can be inserted into 2D CNNs to enable spatial-temporal feature learning.}
\label{fig:tsm}
\vspace{-10pt}
\end{figure}

\begin{figure}[t]
\centering
\begin{subfigure}[b]{0.3\textwidth}
        \centering
        \includegraphics[width=\linewidth]{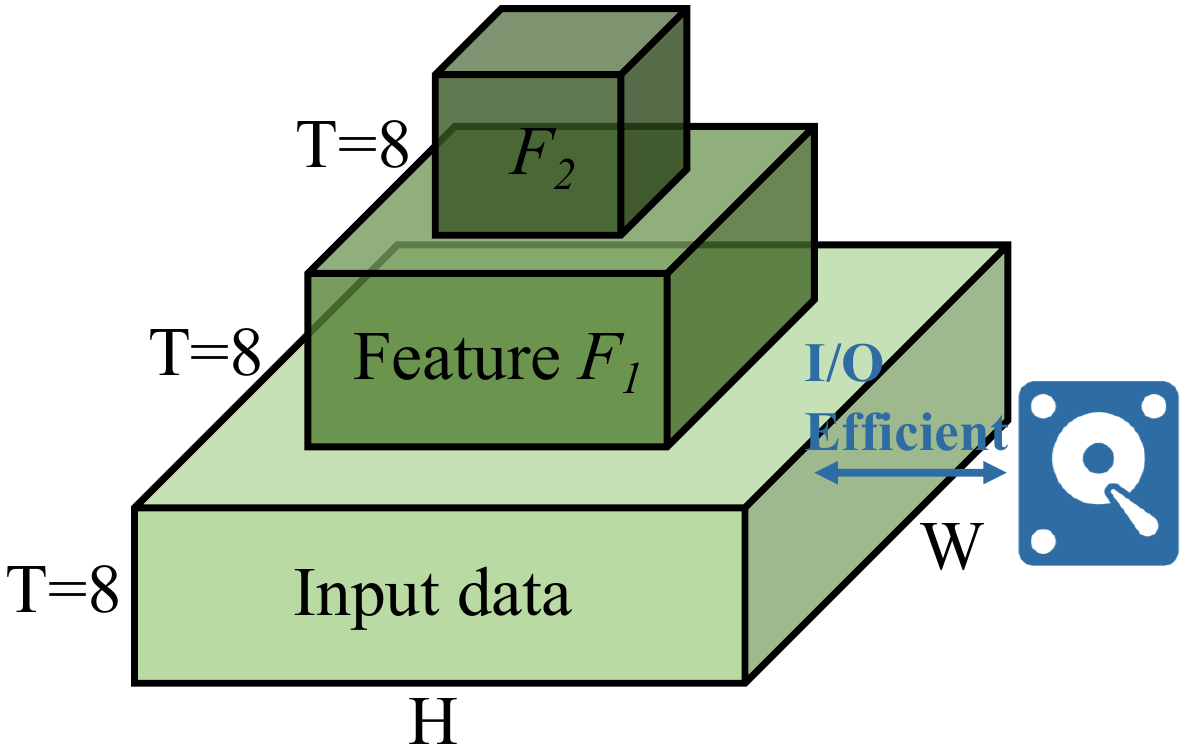}
        \caption{Straight-up}
        \label{fig:straightup}
    \end{subfigure}%
    ~~
    \begin{subfigure}[b]{0.31875\textwidth}
        \centering
        \includegraphics[width=\linewidth]{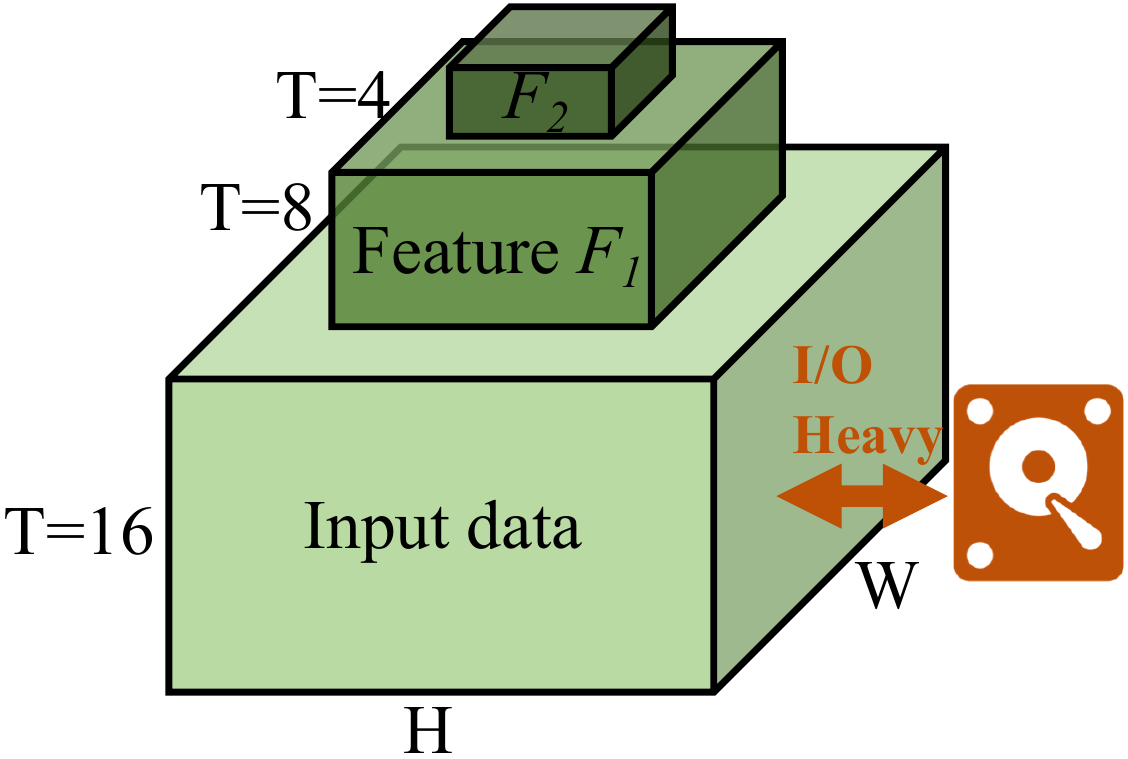}
        \caption{Pooled-up}
        \label{fig:pooledup}
    \end{subfigure}
\caption{Two kinds of video backbone design. Straight-up backbone does not perform temporal pooling and is more data efficient. Pooled-up version requires many input frames and drains I/O.}
\label{fig:test2}
\vspace{-15pt}
\end{figure}


\myparagraph{Temporal modeling unit.} 
3D convolution is the most widely used operator for spatial-temporal modeling. However, it suffers from two problems: (1) large computation and large parameter size, which slows down training and communication; (2) low hardware efficiency compared to 2D convolution (see Section~\ref{sec:guidelines} for details). 2D convolution kernels are highly optimized on various hardware and systems~\cite{chetlur2014cudnn, chen2018tvm}, but 3D convolutions are less optimized. Give the same amount of FLOPs, 3D kernels run 1.2 to 3 times slower than 2D on cuDNN~\cite{chetlur2014cudnn}. To deal with these issues, a more efficient operator is the Temporal Shift Module (TSM)~\cite{lin2018tsm}. As shown in Figure~\ref{fig:tsm}, TSM module shifts part of the channels along temporal dimension to mingle the feature between consecutive frames. In this way, we can perform temporal feature learning without incurring extra computation cost. With TSM, we can build a spatial-temporal video network using pure 2D CNN, enjoying high hardware efficiency.

\myparagraph{Backbone topology.}
Existing video networks usually sample \textbf{many} frames as input (32 frames~\cite{wang2017non} or 64 frames~\cite{carreira2017quo}), and perform temporal pooling later to progressively \textit{reduce} the temporal resolution (Figure~\ref{fig:pooledup}). Another way is to sample \textbf{fewer} frames (e.g. 8 frames~\cite{lin2018tsm}) as input while keeping the \textit{same} temporal resolution to keep the information (Figure~\ref{fig:straightup}). Although the overall computation of the two designs are similar, the former significantly increases the data loading traffic, making the system I/O heavy, which could be quite challenging in a distributed system considering the limited disk bandwidth. 

\section{Design Guidelines to Video Model Architecture} \label{sec:guidelines}

\renewcommand \arraystretch{0.9}
\begin{table*}[t]
\caption{Efficiency statistics of different models. Arrows show the better direction.} 
\vspace{-10pt}
\label{tab:compared_model}
\small
\begin{center}
\begin{tabular}{lcccccc}
\toprule
& \textbf{Acc.$\uparrow$} &  \textbf{FLOPs$\downarrow$} &  \textbf{\#Param.$\downarrow$} &  \textbf{Input size$\downarrow$} & \textbf{Throughput$\uparrow$} & \textbf{Compute/IO$\uparrow$}\\  
\midrule
\itddd~\cite{hara2018can} & 68.0\% & 40G & 47.0M & 16$\times$3$\times$224\textsuperscript{2} & 63.1V/s (1.5$\times$) & 16.6k (2.4$\times$)\\
\itd~\cite{wang2017non} & 73.3\% & 33G & 29.3M & 32$\times$3$\times$224\textsuperscript{2} & 41.9V/s (1.0$\times$) & 6.85k (1$\times$)\\
\midrule
TSM~\cite{lin2018tsm} & \textbf{74.1\%} & 33G & 24.3M & 8$\times$3$\times$224\textsuperscript{2} & 84.8V/s (\textbf{2.0$\times$}) & 27.4k (\textbf{4$\times$}) \\
\bottomrule 
\end{tabular}
\end{center}
\end{table*}

\begin{figure}[t]
\centering
\begin{subfigure}[b]{0.3\textwidth}
        \centering
        \includegraphics[height=1.4in]{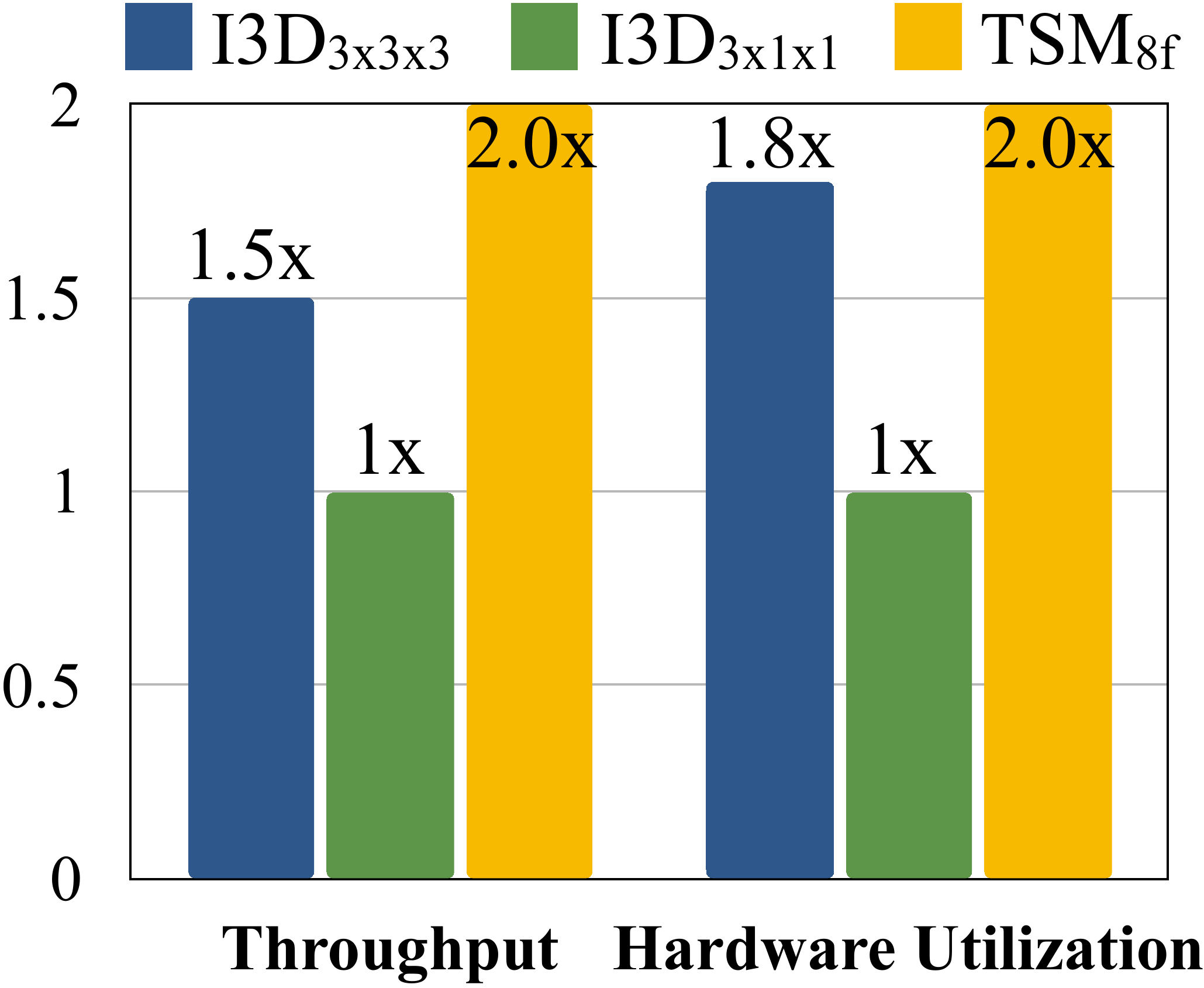}
        \caption{TSM has fewer FLOPs, better throughput and utilization.}
        \label{fig:hardware_efficiency}
    \end{subfigure}%
    ~ 
    \begin{subfigure}[b]{0.3\textwidth}
        \centering
        \includegraphics[height=1.3in]{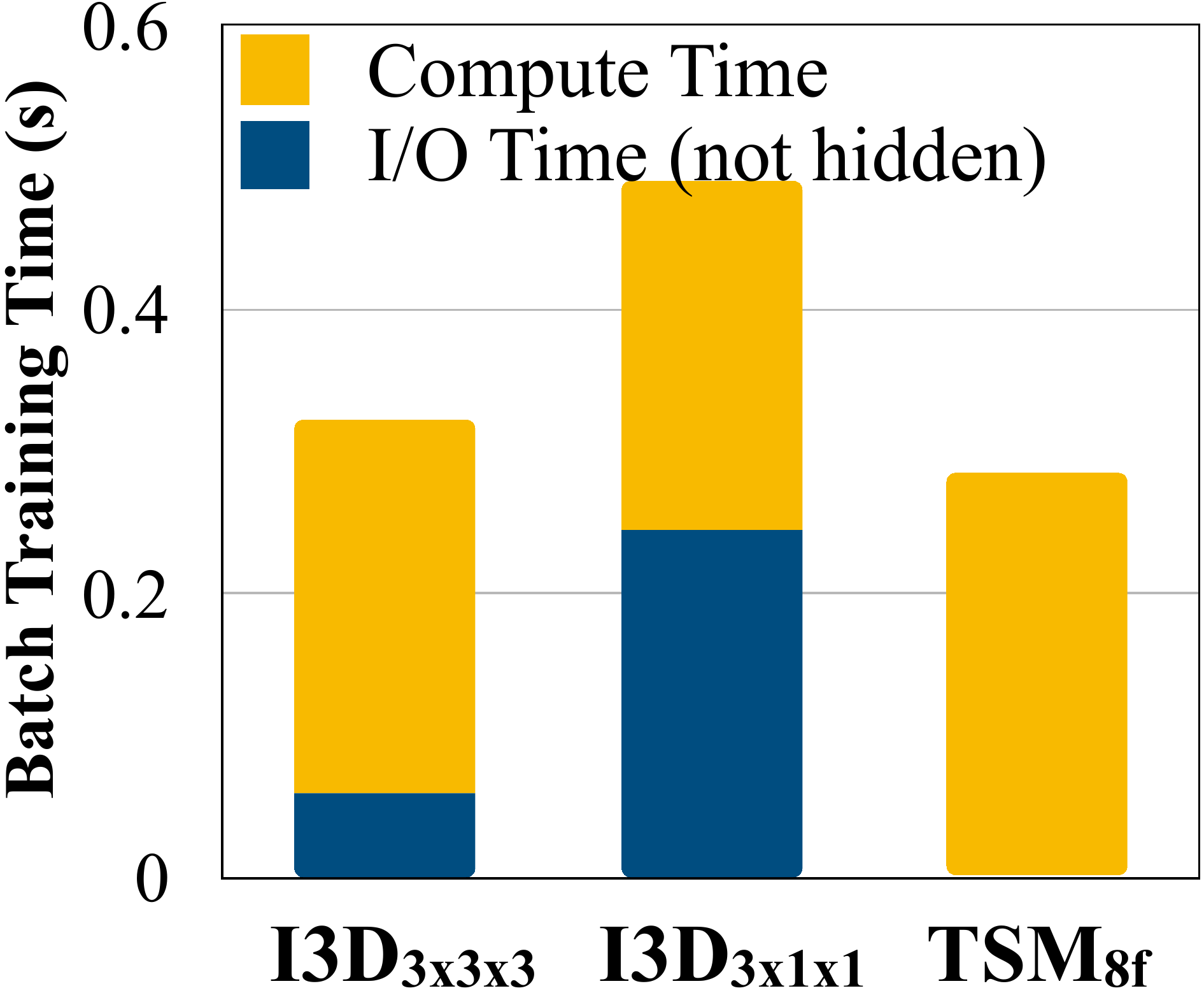}
        \caption{TSM is I/O light, decreasing the total batch time.}
        \label{fig:batch_timing}
    \end{subfigure}
     ~ 
    \begin{subfigure}[b]{0.3\textwidth}
        \centering
        \includegraphics[height=1.3in]{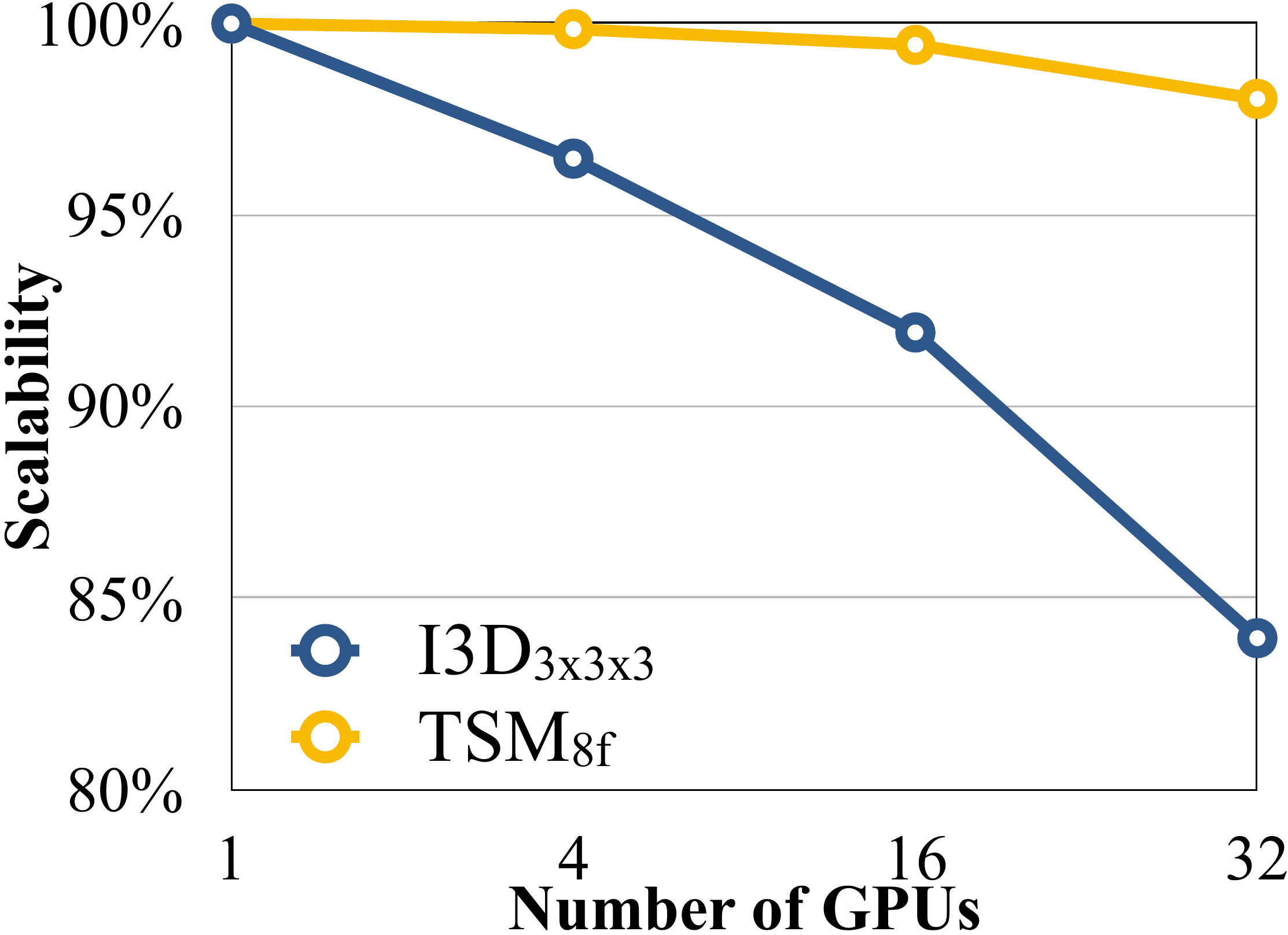}
        \caption{TSM has better scalability due to smaller model size.}
        \label{fig:model_scalability}
    \end{subfigure}
\caption{Analyzing how different design aspects influence the distributed training scalability of video recognition models: \textbf{(a) computation efficiency; (b) data loading efficiency; (c) networking efficiency}.}
\label{fig:guideline}
\end{figure}

To tackle the challenge in a distributed training systems, we propose three video model design guidelines: (1) To increase the \emph{computation efficiency}, use operators with lower FLOPs and higher hardware efficiency; (2) To reduce \emph{data loading traffic}, use a network topology with higher FLOPs/data ratio; (3) To reduce the \emph{networking traffic}, use operators with fewer parameters.

We show the advantage of the above three design guidelines by experimenting on three models in Table~\ref{tab:compared_model}. All the models use the ResNet-50 backbone to exclude the influence of spatial modeling. 
The model architectures are introduced as follows.

(1) The first model is an I3D model from~\cite{hara2018can}. The model takes 16 frames as input and inflate all the $3\times3$ convolutions to $3\times3\times3$. It performs temporal dimension pooling by four times to reduce the temporal resolution. We denote the model as \itddd. 

(2) The second model is an I3D model from~\cite{wang2017non}, taking 32 frames as input and inflating the first $1\times1$ convolution in every other ResBlock. It applies temporal dimension pooling by three times.
We denote this more computation and parameter efficient design as \itd. 

(3) The third model is built with TSM~\cite{lin2018tsm}. The TSM operator is inserted into every ResBlock. The model takes 8 frames as input and performs no temporal pooling. We denote this model as TSM. 

\renewcommand \arraystretch{1.1}
\begin{table}[t]
\caption{The temporal resolution of output feature map for each block. 
TSM is a fully 2D structure, enjoying the best hardware efficiency. The last several stages of \itddd have fewer temporal resolution, making it more similar to 2D CNN, thus enjoying better hardware efficiency compared to \itd. }
\label{tab:network_structure}
\small
\begin{center}
\begin{tabular}{l|c|ccccccc|c}
\textbf{Block} & data & conv\textsubscript{1} & pool\textsubscript{1} & res\textsubscript{2} &  res\textsubscript{3} &  res\textsubscript{3} & res\textsubscript{4} &  res\textsubscript{5} & global pool \\
\hline
\textbf{\itddd} & 16 & 16 & 8 & 8 & - & 4 & 2 & 1 & 1 \\
\textbf{\itd} & 32 & 16 & 8 & 8 & 4 & 4 & 4 & 4 & 1 \\ \hline
\textbf{\tsm} & 8 & 8 & 8 & 8 & - & 8 & 8 & 8 & 1 \\
\end{tabular} 
\end{center}
\vspace{-15pt}
\end{table}

\myparagraph{a. Computation Efficiency.} Computation efficiency is the most direct factor that influence the training time. As shown in Table~\ref{tab:compared_model}, TSM\textsubscript{8f} has 1.2$\times$ fewer FLOPs compared to \itddd and roughly the same FLOPs compared to \itd. However, the actual inference throughput also depends on the hardware utilization. We measure the inference throughput (defined as videos per second) of the three models on a single NVIDIA Tesla P100 GPU using batch size 16. We also measured the hardware utilization, defined as achieved FLOPs/second over peak FLOPs/second. The inference throughput and the hardware efficiency comparison is shown in Figure~\ref{fig:hardware_efficiency}. We can find that the \emph{model is more hardware efficient if it has more 2D convolutions than 3D}: TSM is a fully 2D CNN, therefore it has the best hardware utilization (2.0$\times$); while the last several stage of \itddd (res\textsubscript{4}, res\textsubscript{5}) have few temporal resolution (as shown in Table~\ref{tab:network_structure}), it is more similar to 2D convolution and thus is 1.8$\times$ more hardware efficient than \itd  (1.0$\times$).


\myparagraph{b. Data Loading Efficiency.} Video datasets are usually very large. For a distributed system like the Summit supercomputer, the data is usually stored in High Performance Storage System (HPSS) shared across all the worker nodes. Such file systems usually have great sequential I/O performance but inferior random access performance. Therefore, large data traffic could easily become the system bottleneck. Previous popular I3D models~\cite{carreira2017quo, hara2018can} takes many frames per video (16 or 32) as input and perform down-sample over temporal dimension. We argue that such design is a waste of disk bandwidth: a TSM\textsubscript{8f} only takes 8 frames as input while achieving better accuracy. The intuition is that nearby frames are similar; loading too many similar frames is redundant. 
We empirically test the data loading bottleneck on Summit. To exclude the communication cost from the experiments, we perform timing on single-node training. We measure the total time of one-batch training and the time for data loading (that is not hidden by the computation). As shown in Figure~\ref{fig:batch_timing}, for \itd, it takes 32 frames as input. The data loading time cannot be hidden by the computation, therefore data I/O becomes the bottleneck. \itddd that takes 16 frame as input has less problem on data loading, while TSM\textsubscript{8f} can fully hide the data loading time with computation. We also compute the model FLOPs divided by the input data size as a measurement of data efficiency. The value is denoted as "Compute/IO" as in Table~\ref{tab:compared_model}. For scalable video recognition models, we want a model with larger Compute/IO ratio.

\myparagraph{c. Networking Efficiency.} In distributed training system, the communication time can be modelled as:
\begin{equation}
    \text{communication time} =  \text{latency} + \frac{\text{model size}}{\text{bandwidth}}
\end{equation}
The latency and bandwidth is determined by the network condition, which cannot be optimized through model design. However, we can reduce the model size to reduce the communication cost. Both I3D models inflate some of the 2D convolution kernels to 3D, which will increase the number of parameters by $k_T$. While TSM module does not introduce extra parameters. Therefore, it has the same model size as the 2D counterpart. For example, \itddd has 1.9$\times$ larger model size than TSM\textsubscript{8f}, which introduces almost two times of network communication during distributed training.
To test the influence of model size on scalability, we measure the scalability on a 8 node cluster. Each computer has 4 NVIDIA TESLA P100 GPUs. We define the scalability as the actual training speed divided by the ideal training speed (single machine training speed * number of nodes). The results are shown in Figure~\ref{fig:model_scalability}. Even with the high-speed connection, the scalability of \itddd quickly drops as the number of training nodes increase: the scalability is smaller than 85\% when applied to 8 nodes. While \tsm model still has over 98\% of scalability thanks to the smaller model size thus smaller networking traffic.

\section{Large-scale Distributed Training on Summit}

In this section, we scale up the training of video recognition model on Summit supercomputer. With the help of above hardware-aware model design techniques, we can scale up the training to 1536 GPUs, finishing the training of Kinetics in 15 minutes.

\subsection{Setups}

\myparagraph{Summit supercomputer.}
Summit~\cite{vazhkudai2018design} or OLCF-4 is a supercomputer at Oak Ridge National Laboratory, which as of September 2019 is the fastest supercomputer in the world. It consists of approximately 4,600 compute nodes, each with two IBM POWER9 processors and six NVIDIA Volta V100 accelerators. The POWER9 processor is connected via dual NVLINK bricks, each capable of a 25GB/s transfer rate in each direction. Nodes contain 512 GB of DDR4 memory for use by the POWER9 processors and 96 GB of High Bandwidth Memory (HBM2) for use by the accelerators~\footnote{\url{https://www.olcf.ornl.gov/for-users/system-user-guides/summit/summit-user-guide}}.

We used PyTorch and Horovod~\cite{sergeev2018horovod} for distributed training. The framework uses ring-allreduce algorithm to perform synchronized SGD. The training is accelerated by CUDA and cuDNN. 
We used NVIDIA Collective Communication Library (NCCL)~\footnote{\url{https://developer.nvidia.com/nccl}} for most of the communication. 

\myparagraph{Dataset.}
In our experiments, we used Kinetics-400 dataset~\cite{kay2017kinetics}. The dataset contains 400 human action classes, with at least 400 videos per classes. 
It contains human-object interactions such as playing instruments, as well as human-human interactions such as shaking hands and hugging.
It is widely used to benchmark the performance of video recognition models. 

The dataset has roughly 240k training videos and 20k validation videos, each lasts around 10 seconds. Such large scale raises a critical challenge for model training and data storage.

\myparagraph{Training/Testing Setting.}
We follow the commonly used distributed training and testing setting in~\cite{wang2017non}. For training, we train the network for 100 epochs. We denote the number of total GPUs used for training as $k$, and the batch size per GPU is denoted as $n$. The total batch size is $kn$. In our experiments, we trained a TSM network with 8-frame input and used a fixed $n=8$.
The initial learning rate is set to 0.00125 for every 8 samples, and we apply linear scaling rule~\cite{goyal2017accurate} to scale up the learning rate with larger batch size. The total learning rate is $\eta=0.00125k$.
We used cosine learning rate decay with 5 epochs of warm-up~\cite{goyal2017accurate}. We used weight decay 1e-4 and no dropout for training. We applied no weight decay on BatchNorm and bias following~\cite{wang2016temporal, jia2018highly}.

For testing, we follow~\cite{wang2017non} to sample 10 clips per video and calculate the average prediction. The video is spatially resized with a shorter size 256 and fed into network. For the reported results, we calculate the average and standard deviation from the checkpoint of last 5 epochs to reduce the influence of randomness,


\begin{figure}[t]
\centering
\includegraphics[width=0.55\textwidth]{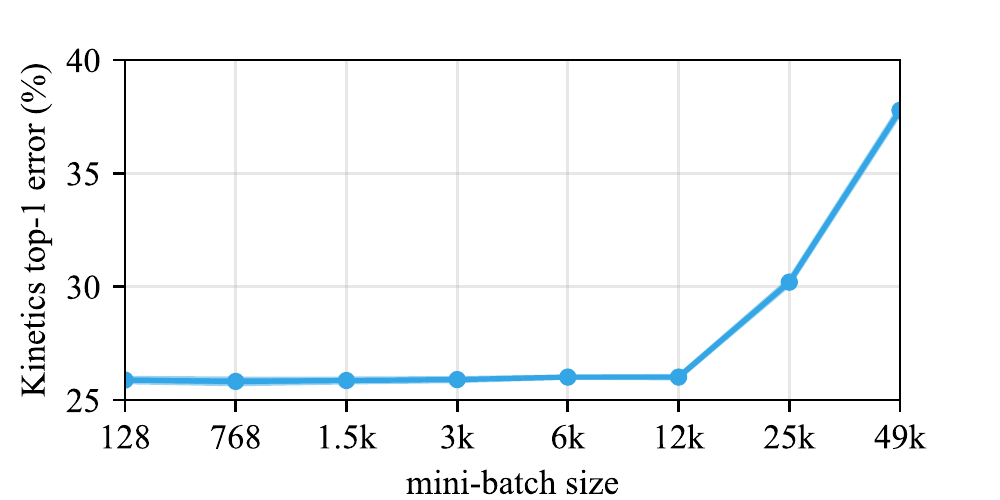}
\caption{Kinetics top-1 validation accuracy \vs mini-batch size. The performance of the model does not degrade when we scale up the mini-batch size to $12k$. The mean and standard deviation (the scale of the STD is hardly visible) are shown in the figure.}
\label{fig:err_vs_bs}
\vspace{-15pt}
\end{figure}
\renewcommand \arraystretch{1}
\begin{table}[t]
\caption{Detailed statistics of different mini-batch size (* indicates simulated performance). }
\label{tab:err_vs_bs}
\small
\begin{center}
\begin{tabular}{lllllll}
\toprule
\textbf{\#Node} &  \textbf{\#GPU} &  \textbf{BatchSize} & \textbf{\#Frames} &  \textbf{Accuracy} &  \textbf{Train Time} & \textbf{\emph{Note}}\\
\midrule
1 & 6 & 96 & 768 & 74.12$\pm$0.11 & 49h 55m & \emph{Baseline}\\
\midrule
8 & 48 & 384 & 3,072 & 74.12$\pm$0.08 & 7h 7m & \multirow{6}{*}{\emph{\shortstack{same level \\ of \\ accuracy}}}\\
16 & 96 & 768 & 6,144 & 74.18$\pm$0.14 & 3h 38m &  \\
32 & 192 & 1,536 & 12,288 &  74.14$\pm$0.10 &  1h 50m  \\
64 & 384 & 3,072 & 24,576 & 74.10$\pm$0.08 & 55m 56s\\
128 & 768 & 6,144 & 49,152 & 73.99$\pm$0.04 & 28m 14s\\
256 & 1536 & 12,288 & 98,304 & 73.99$\pm$0.07 & 14m 13s \\
\midrule
384 & 2304 & 18,432 & 147,456 & 72.52$\pm$0.07 & 10m 9s & \multirow{3}{*}{\emph{\shortstack{lose \\ accuracy}}}\\
512* & 3072 & 24,576 & 196,608 & 69.80$\pm$0.13 & - \\
1024* & 6144 & 49,152 & 393,216 & 62.22$\pm$0.17 & -\\
\bottomrule 
\end{tabular}
\end{center}
\vspace{-10pt}
\end{table}

\subsection{Experiments}

\myparagraph{Baseline.}

For the baseline, we trained a ResNet-50 \tsm model on a single Summit node with 6 GPUs, each GPU contains 16 video clips, resulting in a total batch size of $kn=96$. We evaluate the performance of last 5 checkpoints, it achieves a top-1 accuracy of $74.12\pm 0.11\%$.
 
\myparagraph{Performance \vs Batch Size.}

We first compare the training error \vs the batch size. As shown in~\cite{goyal2017accurate}, the accuracy will not degrade when the batch size is relatively small. Therefore, our experiments start from 8 computing nodes (48 GPUs, 384 video clips, 3072 frames) to 1024 computing nodes (6144 GPUs, 49152 video clips, 393216 frames) . Note that each sample in a video recognition model is a video clip consisting of several frames/images (in our case, 8). Therefore, the actual number of images used in one batch could be much larger than ImageNet training (\eg, $98k$ \vs $8k$~\cite{goyal2017accurate}).

\begin{figure*}[t]
\centering
\begin{subfigure}[b]{0.32\textwidth}
	\includegraphics[width=\textwidth]{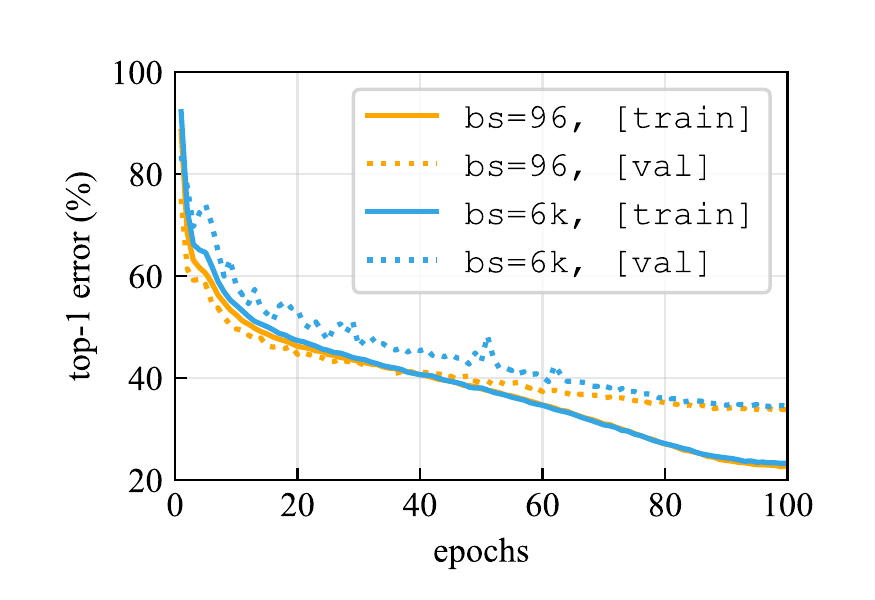}
	\caption{Mini-batch size $6k$.}
	\label{fig:curve6k}
\end{subfigure}
\begin{subfigure}[b]{0.32\textwidth}
	\includegraphics[width=\textwidth]{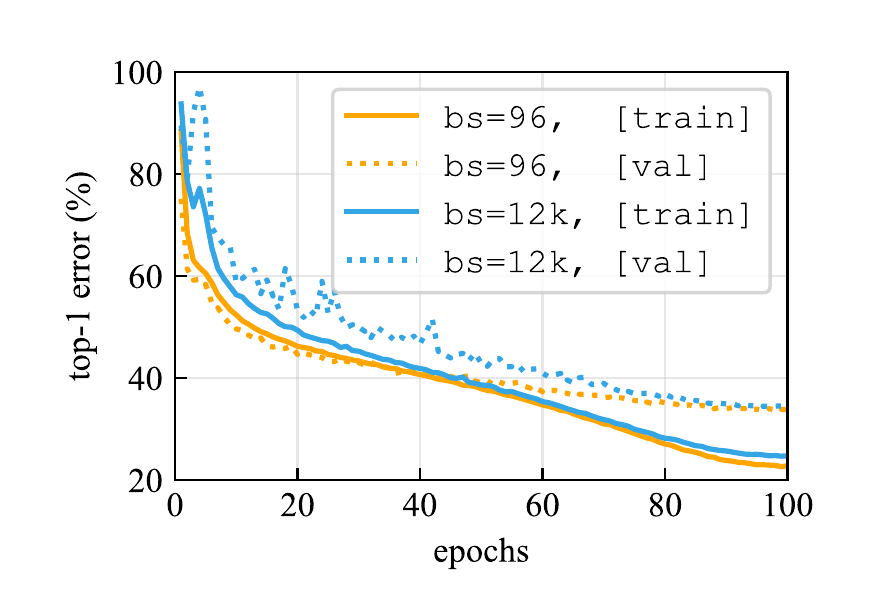}
	\caption{Mini-batch size $12k$.}
	\label{fig:curve12k}
\end{subfigure}
\begin{subfigure}[b]{0.32\textwidth}
	\includegraphics[width=\textwidth]{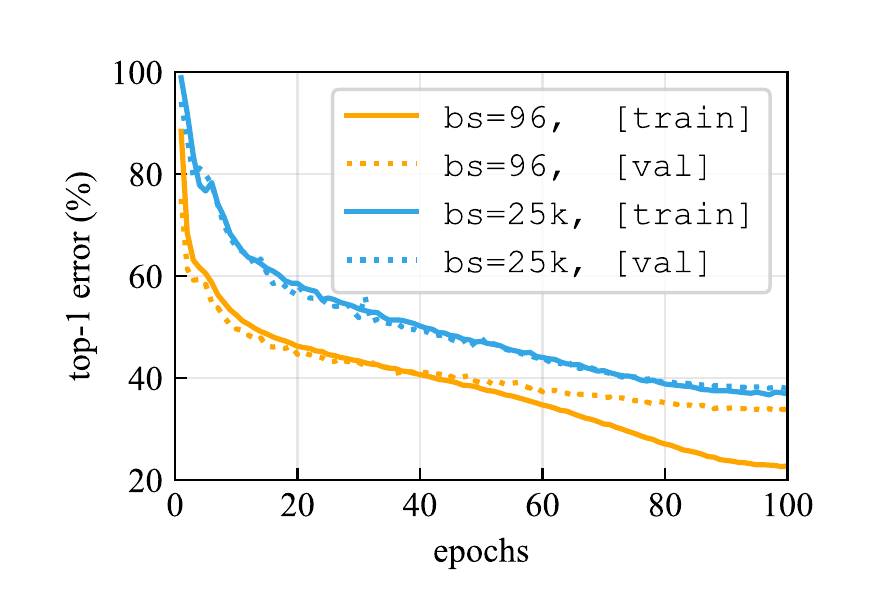}
	\caption{Mini-batch size $25k$ (degrade).}
	\label{fig:curve25k}
\end{subfigure}
\caption{The learning curve for baseline training and large-batch distributed training (batch size 6144, 12228, 24576). The performance does not degrade for batch size 6144 and 12228, while degrades for a batch size of 24576.}
\label{fig:training_curve}
\vspace{-10pt}
\end{figure*}
\begin{figure}[t]
\centering
\includegraphics[width=0.7\textwidth]{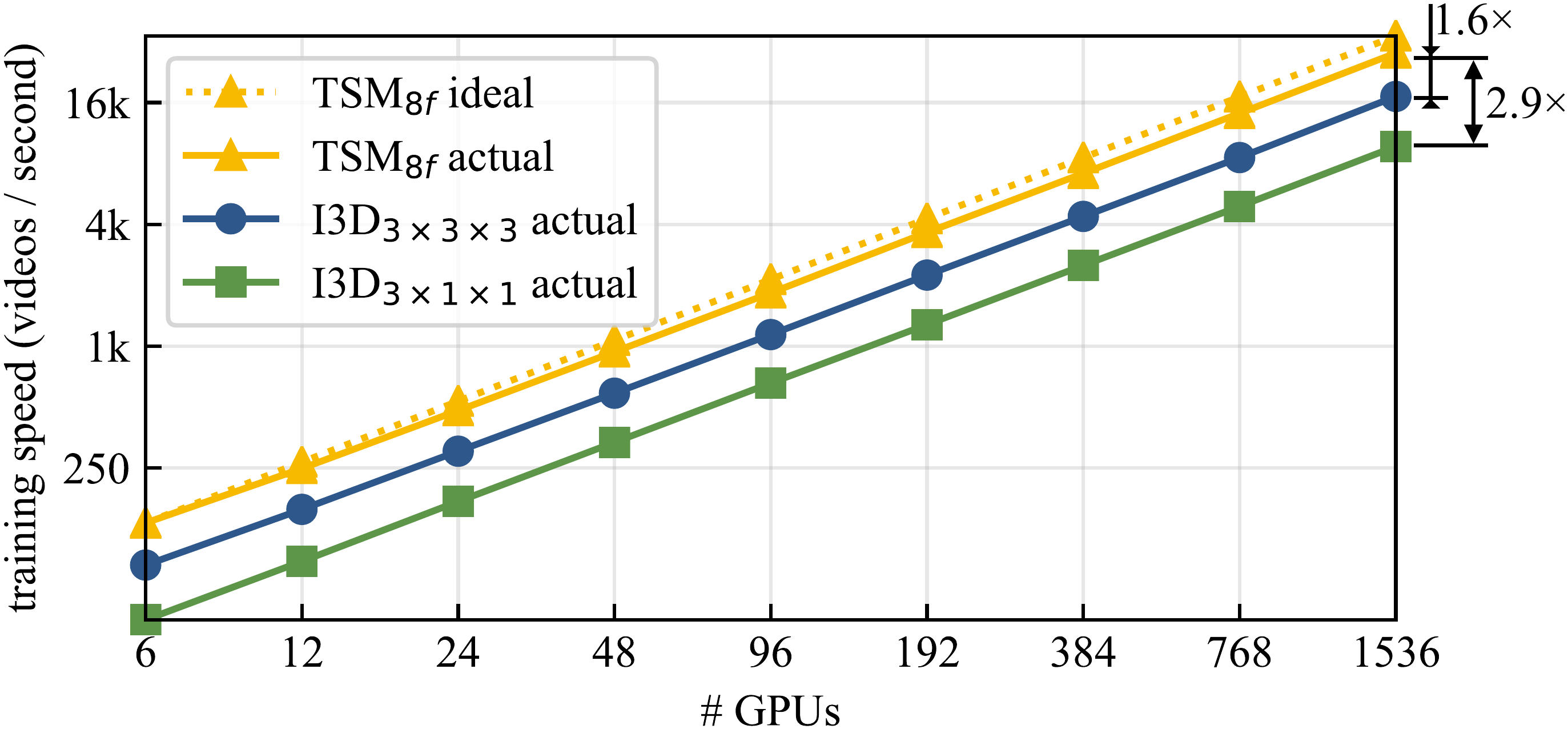}
\caption{The training speed and scalability of distributed synchronous SGD training. \tsm achieves a good scalability ($>$80\%) even when using 1536 GPUs. 
\tsm can achieve 1.6$\times$ higher training speed compared to \itddd and 2.9$\times$ compared to \itd, showing the effectiveness of the proposed design guidelines.}
\label{fig:scalability}
\vspace{-5pt}
\end{figure}

We first plot the error \vs batch size trade-off in Figure~\ref{fig:err_vs_bs}. The error does not increase when we scale the number of computing nodes up to 256 (1536 GPU), where the batch size is 12288, the total frame number is 98304. The detailed statistics are shown in Table~\ref{tab:err_vs_bs}. The scalability of TSM model is very close to the ideal case. Note that due to quota limitation, the largest physical nodes we used is 384 with 2304 GPUs. For 512 and 1024 nodes, we used gradient accumulation to simulate the training process (denoted by *).

We also provide the training and testing convergence curves using 768, 1536 and 3072 GPUs in Figure~\ref{fig:training_curve}. For 768 GPUs and 1536 GPUs, although the convergence of large-batch distributed training is slower than single-machine training baseline, the final converged accuracy is similar, so that the model does not lose accuracy. For 3072 GPUs, the accuracy degrades for both training and testing.


\myparagraph{Scalabilty}

We test the scalability of distributed training on Summit. According to the results from last section, we can keep the accuracy all the way to 256 computing nodes. Therefore, we sweep the number of computing nodes from 1 to 256 to measure the scalability. We keep a batch size of 8 for each GPU and each node has 6 GPUs. So the batch sizes change from 48 to 18,432. Each video clips contains 8 frames in our model, resulting a total number of frames from 384 to 147,456.
We measure the training speed (videos/second) to get the actual speed-up. We calculate the ideal training speed using the single node training speed multiplied by number of nodes. The comparison of different models is provided in Figure~\ref{fig:scalability}. The actual training speed is just marginally below the ideal scaling, achieving $>80\%$ scaling efficiency. 
We also provide the detailed overall training time in Table~\ref{tab:err_vs_bs}.
With 1536 GPUs, we can finish the Kinetics training with TSM within 14 minutes and 13 seconds, achieving a top-1 accuracy of $74.0\%$. The overall training speed of \tsm is 1.6$\times$ larger than \itddd and 2.9$\times$ larger than \itd, showing the advantage of hardware-aware model design.






\section{Conclusion}

In this paper, we analyze how the design of video recognition models will affect the distributed training scalability. We propose three design guidelines to increase \emph{computation efficiency, data loading efficiency, and networking efficiency}. With these guidelines, we designed a special TSM model and scale up the training on Summit supercomputer to 1536 GPUs, achieving 74.0\% top-1 accuracy on Kinetics dataset within 14 minutes and 13 seconds.

\paragraph{Acknowledgments}
We thank John Cohn, Oak Ridge National Lab, MIT Quest for Intelligence, MIT-IBM Watson AI Lab, MIT-SenseTime Alliance, and AWS for supporting this work.

{\small
\bibliographystyle{plain}
\bibliography{main}
}

\end{document}